# Mitigating Nonlinear Algorithmic Bias in Binary Classification

Wendy Hui
Infocomm Technology Cluster
Singare Institute of Technology
Singapore
wendy.hui@singaporetech.edu.sg

Wai Kwong Lau
School of Physics, Mathematics and Computing
University of Western Australia
Australia
john.lau@uwa.edu.au

*Abstract*— **This paper proposes the use of causal modeling to detect and mitigate algorithmic bias that is nonlinear in the protected attribute. We provide a general overview of our approach. We use the German Credit data set, which is available for download from the UC Irvine Machine Learning Repository, to develop (1) a prediction model, which is treated as a black box, and (2) a causal model for bias mitigation. In this paper, we focus on age bias and the problem of binary classification. We show that the probability of getting correctly classified as "low risk" is lowest among young people. The probability increases with age nonlinearly. To incorporate the nonlinearity into the causal model, we introduce a higher order polynomial term. Based on the fitted causal model, the de-biased probability estimates are computed, showing improved fairness with little impact on overall classification accuracy. Causal modeling is intuitive and, hence, its use can enhance explicability and promotes trust among different stakeholders of AI.**

## I. INTRODUCTION

AI applications in areas such as healthcare, education, and fraud investigation often have ethical implications. Beyond ensuring the utility of AI applications, it is essential to address their trustworthiness. The European Commission [1] has outlined four key ethical principles for trustworthy AI, namely, (i) respect for human autonomy, (ii) prevention of harm, (iii) fairness, and (iv) explicability. This paper focuses the principle of *fairness*, which requires that AI systems should avoid unfair bias, promote diversity, and guarantee accessibility for users with diverse abilities. We propose the use of causal modelling to provide post-processing statistical remedies to mitigate algorithmic bias. Our approach makes use of statistical techniques that are easily interpretable, thereby enhancing *explicability* and promoting trust among different stakeholders.

In the context of AI fairness, sensitive data such as gender, race, age, religion, etc., are known as *protected attributes* [2]. The goal of AI fairness is to ensure that outcomes from AI do not exhibit any bias based on these attributes. However, fairness can be defined and measured differently. One approach to establish fairness is by *fairness by unawareness*, which means that machine learning algorithms exclude protected attributes in the training process. However, this approach ignores the fact that other features, such as a person's hobby or address, may correlate with the protected attributes, resulting algorithmic bias even after removing the protected attributes from training. Another approach to fairness is to enforce *demographic parity*, which requires AI output to maintain an acceptable level of disparity between protected and non-protected groups. However, demographic parity may lead to "reverse discrimination." For example, in job hiring, it could result in the acceptance of unqualified individuals in the protected group in order to ensure demographic disparity [3]. Consider a scenario where a protected attribute is correlated with education, which is used to predict the hiring outcome. It is not surprising that the prediction would be correlated with the protected attribute. Makhlouf et al. [4] consider this type of demographic disparity to be justifiable and recommends the use of causal modelling to account for the disparity.

This paper aims to contribute to our understanding of AI fairness by applying causal modelling in the detection and mitigation of bias introduced by an algorithm. We will focus on the problem of binary classification. While previous works mitigated bias in terms of categorical protected attributes [3, 5], we will present a case of algorithmic bias in terms of numeric protected attributes. Furthermore, we will demonstrate how causal modelling can be used to mitigate algorithmic bias that is nonlinear in the protected attributes.

In the remainder of the paper, we will review some related work, describe our method, present our analysis, evaluate the effectiveness of our method, discuss our results, summarize the paper, and identify a number of future research directions.

## II. RELATED WORK

There are three general approaches to mitigating algorithmic bias: (1) pre-processing, (2) in-processing and (3) post-processing. Pre-processing involves manipulating data before training the AI model. In-processing incorporates fairness in the learning algorithm. For example, Celis et al. [6] incorporate fairness constraints for a large family of classification problems. Post-processing aims to reduce bias by manipulating the output from AI. It separates model training from bias mitigation, allowing practitioners to use existing learning algorithms without any modification. In the bias mitigation phase, the trained model can be treated as a black box without the need for re-training. Manipulation of predictions from the trained model can be achieved using standard statistical packages [5]. Thus, post-processing does not require the customization of processes for specific data sets and this advantage can represent significant time saving when large volumes of data are involved. Our method belongs to the category of post-processing.

Our method involves the use of causal modelling, which is relatively new in AI. Most of the works in this area are based on [7] and [8]. Causal models allow us to determine what a prediction would be like in an alternate universe where there is no bias. Kusner et al. [9] defined the concept of *counterfactual fairness*, which requires a decision to remain the same in a counterfactual world where the individual concerned belongs to a different demographic group. Using a health data set, Madras et al. [10] showed that confounding factors can be incorporated into causal models to improve prediction accuracy. Khademi et al. [11] demonstrated the use of causal modelling to detect biases using a synthesized data set and two publicly available data sets.

In addition to bias detection, [3] and [12] also mitigated the detected bias. Focusing on binary classification, [3] demonstrates how fairness can be achieved by equalized odds or equal opportunity, and thus allowing the protected attributes to correlate with the target. [12] extended [3] to provide the solution for establishing multiclass fairness. [5] demonstrated that bias mitigation can be easily achieved using the "lavaan" package in R. However, [3], [5], and [12] only analysed protected attributes that are categorical. This paper builds on [5] but focuses on a numeric protected attribute (age) that exhibits a nonlinear relationship with algorithmic bias.

## III. METHOD

Causal modelling is an extension of regression and allows for more than one endogenous (i.e., dependent) variable in the model. The technique also allows for the incorporation of unobservable, latent variables [13]. As discussed in [14], it is important to define the causal model properly; otherwise, results from the analysis could be misleading. As we focus only on bias that is introduced by the algorithm (instead of bias that already exist in the training data), our causal model follows that of [5] and is graphically presented in Fig. 1.

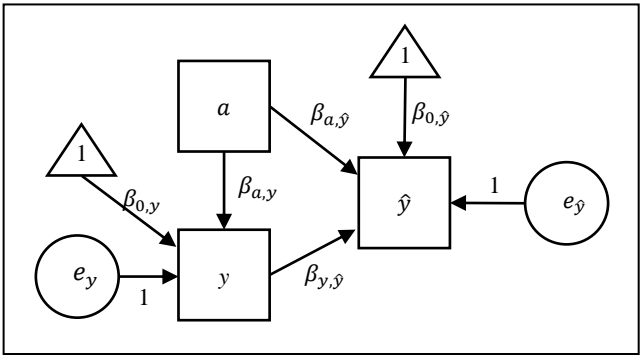

Fig. 1. The Causal Model [5]

Fig. 1 follows the path diagram convention presented in [15], where observable variables are depicted as rectangles. In this case, the observable variables are the protected attribute of interest ($a$), the target variable ($y$) and the predictions from the trained model ($\hat{y}$). The trained model is treated as a black box, so we do not need to worry about the learning algorithm involved in the model development. The target variable ($y$) can be categorical (for classification problems) or numeric (for regression problems). Since classification models often produce probability estimates or real-valued scores as output, we assume $\hat{y}$ to be continuous.

Latent (i.e., unobservable) variables are depicted in path diagrams as circles or ovals. In this case, we have the error terms $e_y$ and $e_{\hat{y}}$ as the latent variables. A constant term is represented as a triangle. Here, we use the constant terms "one" to model the intercept terms. The arrows represent directional causal relationships. Fig. 1 suggests that $a$ affects $\hat{y}$, $a$ affects $y$, and $y$ affects $\hat{y}$. Each arrow is associated with a path coefficient that indicates the strength of the causal relationship. Taken together, the path diagram can be translated to the following set of the mathematical equations:

$$\begin{cases} \hat{y} = \beta_{0,\hat{y}} + \beta_{a,\hat{y}}a + \beta_{y,\hat{y}}y + e_{\hat{y}} & (1) \\ y = \beta_{0,y} + \beta_{a,y}a + e_y & (2) \end{cases}$$

Existing causal modelling techniques can be applied to estimate the intercept terms ($\beta_{0,\hat{y}}$ and $\beta_{0,y}$) and the path coefficients ($\beta_{a,y}$, $\beta_{y,\hat{y}}$, and $\beta_{a,y}$). As explained in [15], in the absence of algorithmic bias, there should be no direct relationship between $a$ and $\hat{y}$, because all effects of $a$ should be mediated through $y$. Hence, the coefficient of the path from $a$ to $\hat{y}$ ($\beta_{a,\hat{y}}$) should be zero. If not, based on the developed causal model, we can estimate what $\hat{y}$ should be in the absence of the bias. We denote the de-biased value of $\hat{y}$ as $\tilde{y}$. Decisions based on $\tilde{y}$ will demonstrate improved algorithmic fairness, as we shall show in the following section.

## IV. ANALYSIS

In this section, we will first describe the data set. Next, using the training data, we will develop the prediction model, evaluate the age bias introduced by the prediction model, and illustrate how causal modelling can be used to mitigate the age bias. Finally, we will evaluate the effectiveness of our method using the test data.

We will use R to create the prediction model, create the causal model, and evaluate the bias mitigation based on the causal model. The R script we use for this study is available on Google CoLab[1].

### A. Data Set

The data set we use in this study is the German Credit data set downloadable from the UC Irvine Machine Learning Repository [16]. The data set consists of records of 1,000 individuals. There are 20 features including age (which is our protected attribute ($a$)), number of dependents, account status, credit amount, credit history, etc. The target variable ($y$) is whether or not an individual is considered "high risk." We allocate 70% of the data for training and the remaining 30% for testing. We will analyze the protected attribute age, which is numeric and ranges from 19 to 75.

### B. Training Data

In this subsection, we will use the training data to develop and evaluate (1) the prediction model and (2) the bias mitigation model based on causal modelling.

*1) The Prediction Model:* We implement logistic regression using the "multinom" function in the neural network package "nnet" in R. Other learning algorithms can be used, but predicted probabilities may need calibration [17]. Since the prediction model is treated only as a black box, we will not present the details of the model here.

To generate $\hat{y}$, we use the "predict" function, which yields a probability estimate of an individual's risk level. Dividing the sorted probability estimates into 10 equal groups and comparing the average probability estimate of "high risk" with the actual proportion of "high risk" in each group, Fig. 2 presents the calibration curve (or *reliability diagram* [18]) for the prediction model. It shows that the predictions actually match the observed data quite well.

---
[1] https://colab.research.google.com/drive/16tpLOxfWkCUgywmNhuHEmkYuMXDZ8gmt?usp=sharing

Since 70% of the individuals belong to the "low risk" class, we use the 70th percentile of the prediction $\hat{y}$, which equals to 0.4174, as the threshold for classification. Table I presents the results of the classification based on the training data. To show the nonlinear relationship between age and the algorithmic bias, we have divided the range of age into three roughly equal classes. The "young" class is aged from 19 to 37; the "middle" class is aged from 38 to 56; and the "senior" class is aged from 57 to 75. The numbers in boldface represent correct classifications. Based on Table 1, the classification accuracy is 0.7829.

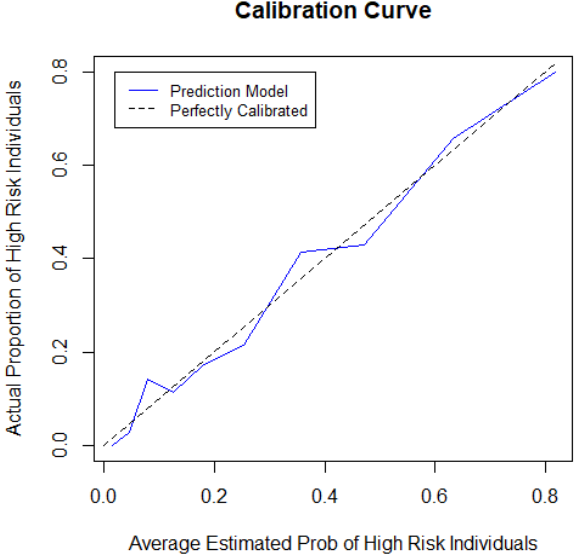

Fig. 2. Calibration Curve for Prediction Model

TABLE I. CLASSIFICATION BASED ON $\hat{y}$ (TRAINING)

|  |  | $\hat{y}$ | |
|---|---|---|---|
| a | y | Low risk (0) | High risk (1) |
| Young (19-37) | Low risk | **245** | 60 |
| Middle (38-56) | Low risk | **132** | 14 |
| Senior (57-75) | Low risk | **39** | 2 |
| Young (19-37) | High risk | 53 | **97** |
| Middle (38-56) | High risk | 17 | **28** |
| Senior (57-75) | High risk | 6 | **7** |

Since there may be justifiable reasons for age to correlate with credit risk (e.g., young people may be less likely to own a home and home ownership is a good indicator of credit risk), it may be unreasonable to require the prediction model to demonstrate demographic parity, i.e., all age groups having the same probability of getting classified as "low risk". Therefore, in this case, we consider algorithmic bias in terms of equal opportunity [3]. Assuming "low risk" is the desirable outcome, the equal opportunity criterion requires that the probability of getting classified as "low risk" given an individual is truly "low risk" is the same across all groups.

Table II evaluates the equal opportunity based on $\hat{y}$. As shown, older individuals tend to have a higher probability of being correctly classified as "low risk," suggesting some degree of algorithmic bias in terms of age. Fig. 2 is a plot of the information in Table II. It shows that the algorithmic bias in terms of equal opportunity is nonlinear.

*2) The Bias Mitigation Model Based on Causal Modeling*: Most statistical tools for analyzing causal models assume linear relationships between variables. To incorporate nonlinearity in the causal model, we revised Equation (1) by including a quadratic term. Hence, our revised set of equations for the causal model becomes:

$$\begin{cases} \hat{y} = \beta_{0,\hat{y}} + \beta_{a,\hat{y}}a + \beta_{a^2,\hat{y}}a^2 + \beta_{y,\hat{y}}y + e_{\hat{y}} & (3) \\ y = \beta_{0,y} + \beta_{a,y}a + e_y & (4) \end{cases}$$

We solve for the causal model above using the "lavaan" package in R. Table III presents the path analysis results (for the endogenous variable $\hat{y}$ only (because the results for $y$ are not relevant to bias mitigation).

TABLE II. EVALUATION OF EQUAL OPPORTUNITY BASED ON $\hat{y}$ (TRAINING)

| a | $Pr\{\hat{y} = \text{"low risk"}|A = a, y = \text{"low risk"}\}$ |
|---|---|
| Young (19-37) | 0.8033 |
| Middle (38-56) | 0.9041 |
| Senior (57-75) | 0.9512 |

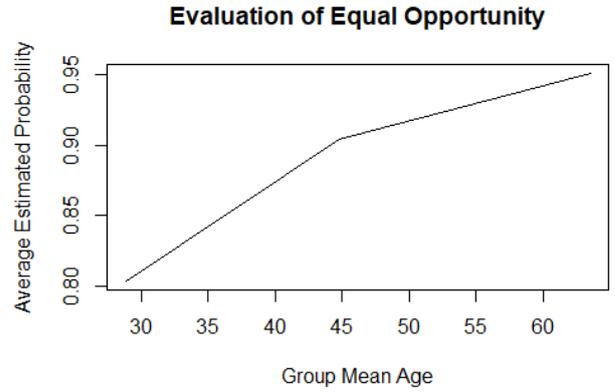

Fig. 3. Comparing Opportunity of Different Age Groups

TABLE III. PATH ANALYSIS RESULTS FOR $\hat{y}$

| Coefficient | | Est. | z-value | p |
|---|---|---|---|---|
| Intercept | $\beta_0$ | 0.4603 | 5.743 | 0.000 |
| Age (a) | $\beta_{a,\hat{y}}$ | -0.0104 | -2.563 | 0.010 |
| Age² (a) | $\beta_{a^2,\hat{y}}$ | $8.3336 \times 10^{-5}$ | 1.751 | 0.080 |
| High risk (y) | $\beta_{y,\hat{y}}$ | 0.3053 | 17.526 | 0.000 |

Since $\beta_{a,\hat{y}}$ is estimated to be significantly different from zero ($z = -2.563$, $p < 0.010$), we can conclude that there is significant age bias introduced by the prediction model, consistent with the results from the equal opportunity analysis in Table 2. Although $\beta_{a^2,\hat{y}}$ is only marginally significant ($z = 1.751$, $p < 0.080$), removing it from the model would lead to an increase of Akaike Information Criterion (AIC) from 690.27 to 691.30. Therefore, we keep it in the model.

Hence, the fitted causal model for $\hat{y}$ is given by:

$$\hat{y} = 0.4603 - 0.0104a + 8.3336 \times 10^{-5}a^2 + 0.3053y.$$

By setting $\beta_{a,\hat{y}}$ and $\beta_{a^2,\hat{y}}$ to zero, we have the de-biased prediction ($\tilde{y}$) as the probability estimate in a counterfactual world where age bias is absent:

$$\tilde{y} = 0.4603 + 0.3053y.$$

With some rearrangement,

$$\tilde{y} = \hat{y} + 0.0104a - 8.3336 \times 10^{-5}a^2. \quad (5)$$

Setting $\beta_{a,\hat{y}}$ and $\beta_{a^2,\hat{y}}$ to zero is the same as assuming everyone's age to be zero when computing $\hat{y}$, thus achieving *counterfactual blindness* to the protected attribute. Therefore, assuming the model specification is correct, the resultant predictions should not be dependent on age. This de-biasing procedure re-ranks the individuals in the data set such that age groups previously favoured by the biased algorithm can no longer enjoy a relative advantage. The de-biased probability estimate $\tilde{y}$ is distributed differently from $\hat{y}$, so the threshold probability for classification needs to be revised. Since 70% of the individuals in the training set belong to the "low risk" class, we set the 70th percentile of $\tilde{y}$ (i.e., 0.6562) as the new classification threshold. Table IV presents the results of the classification based on the training data. Table V evaluates the equal opportunity exhibited by $\tilde{y}$.

TABLE IV. CLASSIFICATION BASED ON $\tilde{y}$ (TRAINING)

| | | $\tilde{y}$ | |
|---|---|---|---|
| *a* | *y* | *Low risk (0)* | *High risk (1)* |
| Young (19-37) | Low risk | **251** | 54 |
| Middle (38-56) | Low risk | **129** | 17 |
| Senior (57-75) | Low risk | **37** | 4 |
| Young (19-37) | High risk | 58 | **92** |
| Middle (38-56) | High risk | 13 | **32** |
| Senior (57-75) | High risk | 5 | **8** |

TABLE V. EVALUATION OF EQUAL OPPORTUNITY BASED ON $\tilde{y}$ (TRAINING)

| *a* | $Pr\{\tilde{y} = "low\ risk" \mid A = a, y = "low\ risk"\}$ |
|---|---|
| Young (19-37) | 0.8230 |
| Middle (38-56) | 0.8836 |
| Senior (57-75) | 0.9024 |

Comparing Table II with Table V, we can see that the numbers presented in Table V are slightly equalized compared to those in Table II, suggesting improved algorithmic fairness in terms of equal opportunity. The classification accuracy after de-biasing is slightly improved from 0.7829 to 0.7843, but the improvement is negligible.

*C. Test Data*

In this subsection, we will use the test data to evaluate the bias mitigation model to ensure that there is no overfitting.

*1) Applying the prediction model:* We apply the prediction model on the test data to compute $\hat{y}$. After that, we apply the same classification threshold identified using the training data for $\hat{y}$ (i.e., 0.4174). Table VI presents the results of the classification. The classification accuracy is 0.7500. As expected, it is worse than that for the biased probability estimation based on the training data. Table VII evaluates the equal opportunity exhibited by $\hat{y}$.

TABLE VI. CLASSIFICATION BASED ON $\hat{y}$ (TEST)

| | | $\hat{y}$ | |
|---|---|---|---|
| *a* | *y* | *Low risk (0)* | *High risk (1)* |
| Young (19-37) | Low risk | **108** | 31 |
| Middle (38-56) | Low risk | **56** | 5 |
| Senior (57-75) | Low risk | **8** | 0 |
| Young (19-37) | High risk | 24 | **38** |
| Middle (38-56) | High risk | 13 | **11** |
| Senior (57-75) | High risk | 2 | **4** |

TABLE VII. EVALUATION OF EQUAL OPPORTUNITY BASED ON $\hat{y}$ (TEST)

| *a* | $Pr\{\tilde{y} = "low\ risk" \mid A = a, y = "low\ risk"\}$ |
|---|---|
| Young (19-37) | 0.7770 |
| Middle (38-56) | 0.9180 |
| Senior (57-75) | 1.0000 |

Like Table II, Table VII shows that young individuals are less likely to be correctly classified as "low risk." The proportion of correct classification for senior, "low risk" individuals seems unrealistically high, but this is because of the small size of this group of individuals (only 8). If we had a larger test data set, the proportion of correct classification for senior, "low risk" individuals would be smaller than 1.

*2) Applying the Bias Mitigation Model:* We apply the same bias mitigation model on the test data to compute $\tilde{y}$. After that, we apply the same classification threshold identified using the training data for $\tilde{y}$ (i.e., 0.6562). Table VIII presents the results of the classification.

TABLE VIII. CLASSIFICATION BASED ON $\tilde{y}$ (TEST)

| | | $\tilde{y}$ | |
|---|---|---|---|
| *a* | *y* | *Low risk (0)* | *High risk (1)* |
| Young (19-37) | Low risk | **110** | 29 |
| Middle (38-56) | Low risk | **53** | 8 |
| Senior (57-75) | Low risk | **8** | 0 |
| Young (19-37) | High risk | 25 | **37** |
| Middle (38-56) | High risk | 12 | **12** |
| Senior (57-75) | High risk | 1 | **5** |

TABLE IX. EVALUATION OF EQUAL OPPORTUNITY BASED ON $\tilde{y}$ (TEST)

| *a* | $Pr\{\tilde{y} = "low\ risk" \mid A = a, y = "low\ risk"\}$ |
|---|---|
| Young (19-37) | 0.7914 |
| Middle (38-56) | 0.8689 |
| Senior (57-75) | 1.0000 |

Comparing Table VII with Table IX, we can see that the numbers presented in Table IX are slightly equalized compared to those in Table VII, suggesting improved algorithmic fairness in terms of equal opportunity. The classification accuracy after de-biasing remains 0.7500.

Table X summarizes the accuracy performance of the classifications based on the biased and de-biased probability estimates.

TABLE X. COMPARISON OF ACCURACY PERFORMANCE

| | | Classification Accuracy | | | |
|---|---|---|---|---|---|
| Data | Estimated Prob. | *Young, Low Risk* | *Middle, Low Risk* | *Senior, Low Risk* | *Overall* |
| Train | Biased $\hat{y}$ | 0.8033 | 0.9041 | 0.9512 | 0.7829 |
| | De-biased $\tilde{y}$ | 0.8230 | 0.8836 | 0.9024 | 0.7843 |
| Test | Biased $\hat{y}$ | 0.7770 | 0.9180 | 1.0000 | 0.7500 |
| | De-biased $\tilde{y}$ | 0.7914 | 0.8689 | 1.0000 | 0.7500 |

V. DISCUSSION

We note that the main objective of this paper is to demonstrate how causal modelling can be used to mitigate nonlinear algorithmic bias. So, there may be more effective ways to handle nonlinearity, e.g., including higher order polynomial terms in the causal model. However, cross-validation using test data should be applied to prevent overfitting.

Discretization of the protected attribute is another way to model the nonlinearity [19]. It involves turning the numeric attribute into a categorical attribute. For this data set, we have created dummy variables for the "middle" and "senior" classes (denoted as $a_m$ and $a_s$) and revised the causal model to:

$$\begin{cases} \hat{y} = \beta_{0,\hat{y}} + \beta_{a_m,\hat{y}} a_m + \beta_{a_s,\hat{y}} a_s + \beta_{y,\hat{y}} y + e_{\hat{y}} & (6) \\ y = \beta_{0,y} + \beta_{a,y} a + e_y & (7) \end{cases}$$

The relevant results of the model fitting are presented in Table XI. The code to generate these results can be found on Google CoLab[2]. As shown, both age classes are significantly different from the "young" class at the 0.05 level.

TABLE XI. PATH ANALYSIS RESULTS FOR $\hat{y}$ (DISCRETIZATION)

| Coefficient | | Est. | z-value | p |
|---|---|---|---|---|
| Intercept | $\beta_0$ | 0.2275 | 19.863 | 0.000 |
| Age = "middle" ($a_m$) | $\beta_{a_m,\hat{y}}$ | -0.0589 | -3.237 | 0.001 |
| Age = "senior" ($a_s$) | $\beta_{a_s,\hat{y}}$ | -0.0902 | -2.961 | 0.003 |
| High risk ($y$) | $\beta_{y,\hat{y}}$ | 0.3118 | 17.788 | 0.000 |

Table XII summarizes the accuracy performance of the classifications based on the biased probability estimates and the probability estimates de-biased using discretization. The statistics again show that the proportion of correct classifications of "young", "low risk" individuals increases with de-biasing, suggesting improved algorithmic fairness in terms of equal opportunity. The changes in the overall classification accuracy are negligible.

TABLE XII. COMPARISON OF ACCURACY PERFORMANCE (DISCRETIZATION)

| Data | Estimated Prob. | Classification Accuracy | | | |
|---|---|---|---|---|---|
| | | Young, Low Risk | Middle, Low Risk | Senior, Low Risk | Overall |
| Train | Biased $\hat{y}$ | 0.8033 | 0.9041 | 0.9512 | 0.7829 |
| | De-biased $\tilde{y}$ | 0.8197 | 0.8904 | 0.9024 | 0.7843 |
| Test | Biased $\hat{y}$ | 0.7770 | 0.9180 | 1.0000 | 0.7500 |
| | De-biased $\tilde{y}$ | 0.7842 | 0.8852 | 1.0000 | 0.7433 |

To summarize, we used causal modelling to detect and mitigate bias in the post-processing stage. We focused on the problem of nonlinear algorithmic bias in binary classification. Our results show that nonlinearity can be incorporated by including higher order polynomial terms or discretization. The de-biased classifications show smaller disparity in terms of equal opportunity, whereas the effect on the overall classification accuracy is negligible.

An advantage of our statistical approach to bias mitigation is *interpretability*. Even though an AI model may be a black box, the post-hoc analysis provides a way to describe the nature of bias and the remedies taken to address the bias. This type of explanation can engender trust among different stakeholders in AI.

There is much room for future research. So far, we have only addressed bias introduced by the prediction model. We will apply causal modelling to address biases that exist within the data. We will also consider other ways biases could be introduced into AI models, including selection bias discussed in [20]. Finally, we will extend our method for multiclass problems.

---

[2] https://colab.research.google.com/drive/13_QqO-5A4hA1c83VC2cbZhudmK0CH1qm?usp=sharing

REFERENCES

[1] European Commission. (2019). "Ethics guidelines for trustworthy AI." Available: https://digital-strategy.ec.europa.eu/en/library/ethics-guidelines-trustworthy-ai

[2] MIT Open Courseware. (2020). "Protected attributes and 'fairness through unawareness." Available: https://ocw.mit.edu/courses/res-ec-001-exploring-fairness-in-machine-learning-for-international-development-spring-2020/pages/module-three-framework/protected-attributes/

[3] M. Hardt, E. Price, and N. Srebro (2019) "Equality of opportunity in supervised learning," Advances in neural information processing systems, vol. 29, pp. 3315–3323, 2016.

[4] K. Makhlouf, S. Zhioua, and C. Palamidessi (2022). "Survey on causal-based machine learning fairness notions." Available: https://arxiv.org/abs/2010.09553

[5] W. Hui and W. K. Lau (2024) "Detecting and mitigating algorithmic bias in binary classification using causal modeling," 4th International Conference on Computer Communication and Information Systems, Phuket, Thailand, February 27-29, 2024. Available: https://doi.org/10.48550/arXiv.2310.12421

[6] L. E. Celis, L. Huang, V. Keswani, and N. K. Vishnoi (2019). "Classification with fairness constraints: A meta-algorithm with provable guarantees." In Proceedings of the 2019 Conference on Fairness, Accountability, and Transparency, Atlanta, GA, USA, pp. 29–31.

[7] J. Pearl (2001). "Direct and indirect effects." In Proceedings of the Seventeenth Conference on Uncertainty in Artificial Intelligence, pp. 411-420.

[8] J. Pearl (2009). Causality. Cambridge University Press.

[9] M. J Kusner, J. Loftus, C. Russell, and R. Silva (2017) "Counterfactual fairness." In Advances in Neural Information Processing Systems, USA, pp. 4066–4076.

[10] D. Madras, E. Creager, T. Pitassi, and R. Zemel "Fairness through causal awareness: Learning causal latent-variable models for biased data," Proceedings of the Conference on Fairness, Accountability, and Transparency, January 2019, pp. 349-358. Available: https://doi.org/10.1145/3287560.3287564

[11] A. Khademi, S. Lee, D. Foley, and V. Honavar (2019) "Fairness in algorithmic decision making: An excursion through the lens of causality," The World Wide Web Conference, May 2019, pp. 2907-2914. Available: https://doi.org/10.1145/3308558.3313559

[12] P. Putzel and S. Lee (2022). "Blackbox post-processing for multiclass fairness." Available: https://arxiv.org/abs/2201.04461

[13] J. M. Youngblut, "A consumer's guide to causal modeling: Part I," Journal of Pediatric Nursing, vol. 9, issue 4, pp. 268-271. Available: https://www.ncbi.nlm.nih.gov/pmc/articles/PMC2905793/

[14] R. Binkyte, K. Makhlouf, C. Pinzon, S. Zhioua, C. Palamidessi (2023) "Causal discovery for fairness." In Proceedings of Machine Learning Research, Vol. 1, pp. 1-16.

[15] Zhang, Z. & Wang, L. (2017). Advanced statistics using R. [https://advstats.psychstat.org]. Granger, IN: ISDSA Press. ISBN: 978-1-946728-01-2.

[16] H. Hofmann. (1994). Statlog (German Credit Data). UCI Machine Learning Repository. Available: https://doi.org/10.24432/C5NC77

[17] Scikit-learn.org (n.d.) "Probability calibration." Available: https://scikit-learn.org/stable/modules/calibration.html

[18] D. S. Wilks (1990) "On the combination of forecast probabilities for consecutive precipitation periods." Weather and Forecasting, Vol. 5, 640–650.

[19] P. Attewell, D. B. Monaghan, and D. Kwong (2015). Data Mining for Social Sciences: An Introduction. University of California Press.

[20] E. Bareinboim and J. Tian (2014). "Recovering causal effects from selection bias." In Proceedings of the Twenty-Ninth AAAI Conference on Artificial Intelligence. Available: https://ftp.cs.ucla.edu/pub/stat_ser/r425.pdf